\documentclass[10pt,twocolumn,letterpaper]{article}

\usepackage{cvpr}              

\usepackage{microtype}
\usepackage[T1]{fontenc}
\usepackage[utf8]{inputenc}
\usepackage[english]{babel}
\IfFileExists{inconsolata.sty}{\usepackage{inconsolata}}{}

\definecolor{cvprblue}{rgb}{0.21,0.49,0.74}
\usepackage[pagebackref,breaklinks,colorlinks,allcolors=cvprblue]{hyperref}

\def\paperID{*****} 
\def\confName{CVPR}
\def\confYear{2026}

\title{Target-Side Paraphrase Augmentation for Sign Language Translation with Large Language Models}

\author{%
\begin{tabular}[t]{@{}c@{\hspace{0.04\textwidth}}c@{}}
\parbox[t][26mm][t]{0.46\textwidth}{\centering
  Pedro Dal Bianco\\
  III-LIDI\\
  Universidad Nacional de La Plata\\
  {\tt\small pdalbianco@lidi.info.unlp.edu.ar}} &
\parbox[t][26mm][t]{0.46\textwidth}{\centering
  Jean Paul Nunes Reinhold\\
  CDTEC\\
  Federal University of Pelotas\\
  {\tt\small jean.pnr@inf.ufpel.edu.br}} \\[-3mm]
\parbox[t][26mm][t]{0.46\textwidth}{\centering
  Oscar Stanchi\\
  III-LIDI\\
  CONICET\\
  {\tt\small ostanchi@lidi.info.unlp.edu.ar}} &
\parbox[t][26mm][t]{0.46\textwidth}{\centering
  Facundo Quiroga\\
  III-LIDI\\
  Comision de Investigaciones Cientificas\\
  Universidad Nacional de La Plata\\
  {\tt\small fquiroga@lidi.info.unlp.edu.ar}} \\[2em]
\parbox[t][26mm][t]{0.46\textwidth}{\centering
  Franco Ronchetti\\
  III-LIDI\\
  Comision de Investigaciones Cientificas\\
  Universidad Nacional de La Plata\\
  {\tt\small fronchetti@lidi.info.unlp.edu.ar}} &
\parbox[t][26mm][t]{0.46\textwidth}{\centering
  Ulisses Brisolara Corrêa\\
  CDTEC\\
  Universidade Federal de Pelotas\\
  {\tt\small ub.correa@inf.ufpel.edu.br}} \\
\end{tabular}%
}

\begin{document}
\maketitle
\begin{abstract}
Sign language translation (SLT) remains constrained by the limited availability of paired sign-video/text corpora and by the heavy-tailed vocabularies typical of real-world datasets. We study a target-side augmentation strategy in which a large language model (LLM) generates controlled paraphrase variants of the reference spoken-language sentence while the sign input remains unchanged. Concretely, we use GPT-4o to produce semantically faithful variants of the training targets and train a Signformer-style pose-based Transformer under a two-stage schedule: pre-training on the augmented corpus followed by fine-tuning on the original references.

We evaluate this strategy on three datasets that span complementary challenges: PHOENIX14T (German Sign Language), a real-world corpus with moderate lexical diversity; the Greek Sign Language Dataset with highly controlled, repetitive recordings; and LSA-T (Argentinian Sign Language), a naturalistic corpus with a large vocabulary and severe long-tail sparsity. This range allows us to characterize precisely when and why target-side augmentation is beneficial.

On PHOENIX14T, augmentation improves BLEU-4 from 9.56 to 10.33, demonstrating that paraphrastic exposure helps the decoder generalize beyond memorized reference phrasing. The near-saturated GSL baseline and the extremely sparse LSA-T setting reveal the limits of the approach: in both cases, single-reference lexical overlap metrics are insufficient to capture the full picture, motivating a complementary semantic evaluation. To our knowledge, this is the first study to examine LLM-generated target-side paraphrases as an augmentation mechanism for SLT, and the first to apply an LLM-as-a-Judge evaluation protocol to SLT. This complementary evaluation reveals gains in semantic fidelity that lexical overlap metrics understate.

\end{abstract}
\section{Introduction}
Sign language translation (SLT) aims to map sign-language video directly to spoken-language text, offering a path toward more accessible communication between deaf signers and hearing communities. SLT sits at the intersection of computer vision and natural language processing, and although the field has advanced substantially in recent years \cite{camgoz2018neural, camgoz2020sign}, robust translation remains difficult.

Signed utterances distribute meaning across manual and non-manual articulators, exhibit strong coarticulation, and vary with signer style and recording conditions. At the same time, large parallel sign-video/text corpora remain scarce \cite{bragg2019sign}, and the datasets that do exist often combine narrow domains, modest sample counts, and heavy-tailed target vocabularies. Many words appear only a handful of times, if at all, making it difficult for translation models to learn stable text-generation behavior. The widely used RWTH-PHOENIX-Weather 2014T (PHOENIX14T) benchmark \cite{camgoz2018neural}, for instance, contains relatively repetitive domain-specific phrasing, whereas broader real-world corpora show much larger vocabularies and many more singletons. The combined effect of multimodal complexity, domain shift, and lexical sparsity remains a central obstacle for SLT.

These constraints also motivate compact, reproducible SLT models. High-capacity video-based systems can deliver stronger absolute performance, but they often depend on costly visual backbones and training pipelines that are difficult to reproduce in smaller research settings. Pose-based models derived from lightweight architectures such as Signformer \cite{yang2024signformer} offer a more practical alternative: by representing each video as a sequence of body and hand landmarks, they reduce input dimensionality and make experimentation more affordable. We adopt that perspective here, using MediaPipe-derived pose features with a Signformer-inspired Transformer so that the effect of data augmentation can be studied within a controlled and resource-efficient setup.

Data augmentation is a standard strategy in low-resource machine translation. In spoken-language MT, techniques such as back-translation and paraphrasing often improve generalization under limited supervision \cite{sennrich2016backtranslation,hu2021textaug}. In sign-language research, however, augmentation has mostly focused on the sign side, either by generating synthetic gloss--text pairs \cite{moryossef-etal-2021-data} or by synthesizing additional sign videos through sign-language production models \cite{walsh2025signaugmentation}. Our focus is different: we augment the \emph{text side} of existing SLT pairs with large language models. More specifically, we study target-side paraphrase augmentation: each reference sentence is rewritten into semantically faithful variants, exposing the decoder to multiple acceptable surface realizations of the same signed content. The goal is to reduce overfitting to rigid reference wording and improve robustness to lexical and syntactic variation at decoding time. To our knowledge, LLM-generated target-side paraphrase augmentation has not previously been studied for SLT, although LLMs have begun to appear in SLT pipelines in other roles, such as stronger text decoders \cite{wong2024sign2gpt}.

We evaluate the method on three datasets spanning different sign languages and levels of difficulty: PHOENIX14T (German Sign Language) \cite{camgoz2018neural}, a weather-forecast corpus with relatively formulaic text; a Greek Sign Language (GSL) educational corpus \cite{adaloglou2020comprehensive}; and an Argentinian Sign Language corpus derived from LSA-T \cite{dalbianco2022lsat}. Our hypothesis is intentionally narrow: target-side augmentation should help the decoder tolerate superficial lexical and syntactic variation while preserving signed meaning, rather than solve broader challenges such as unseen signers or cross-domain transfer.

This also motivates a semantic evaluation alongside BLEU, which we conduct using an LLM-as-a-Judge protocol described in Section~\ref{sec:results}.

Our main contributions are as follows: (1) we introduce LLM-based target-side paraphrase augmentation for sign language translation and release three augmented SLT datasets covering DGS, GSL, and LSA; (2) we present a three-dataset study on PHOENIX14T, GSL, and LSA-T, showing that the effect of this augmentation depends strongly on corpus characteristics, ranging from a BLEU-4 gain on PHOENIX14T to negligible or slightly negative effects in simpler or more extreme long-tail settings; and (3) we conduct a semantic evaluation using an LLM-as-a-Judge protocol on PHOENIX14T and GSL, showing that augmentation yields gains in semantic fidelity that lexical overlap metrics understate. All code and datasets are publicly available \footnote{URL anonymized for review purposes.}.

\section{Related Work}
\paragraph{Sign Language Translation.} Early SLT systems typically adopted a two-stage pipeline: continuous sign language recognition first predicted an intermediate gloss sequence, which was then translated into spoken text \cite{camgoz2018neural}. Glosses provide a convenient intermediate representation, but obtaining them is labor-intensive and they cannot capture the full richness of signed communication, including facial expressions and classifier constructions. More recent work has therefore shifted toward \emph{gloss-free} SLT, learning an end-to-end mapping from video directly to spoken-language text \cite{camgoz2020sign, chen2022twostream}. This setting is more challenging and usually trails gloss-based performance, but it is also more scalable because it requires only video--text pairs. Modern gloss-free systems increasingly rely on transformer architectures and, in some cases, large pretrained models. For example, Sign2GPT \cite{wong2024sign2gpt} combines a pretrained CLIP visual encoder with a GPT-style decoder and lightweight adapters, achieving state-of-the-art results on PHOENIX14T and CSL-Daily. By contrast, \cite{yang2024signformer} proposed \textit{Signformer}, a highly lightweight transformer without pretrained components that still achieved competitive performance. Our model follows that efficiency-oriented line of work, but uses body-pose sequences rather than dense visual embeddings.

\paragraph{Data Augmentation in SLT.} The scarcity of sign-to-text data has encouraged a broad range of augmentation methods. Beyond standard video perturbations such as mirroring or spatial jitter, prior SLT work has explored richer sign-side augmentation strategies. One direction synthesizes additional training examples through sign-language production models \cite{stoll2020text2sign}; another uses motion synthesis, stitching, or generative models such as SignGAN and SignSplat to create new sign video variations, with substantial relative BLEU improvements in some settings \cite{walsh2025signaugmentation}. Target-side augmentation has received less attention in SLT. \cite{moryossef-etal-2021-data} expanded gloss-to-text training data through paraphrase pairs derived from monolingual data and heuristic rules. More broadly in NLP, LLMs such as GPT-3 and GPT-4o have been used to generate paraphrases and synthetic examples for low-resource tasks \cite{davoodi2022lmaug}. Our work brings that idea to SLT by rewriting target-language references while keeping the sign input fixed, making it complementary to sign-side augmentation rather than a substitute for it.

\section{Methodology}

\subsection{Model Architecture}

\begin{figure}[ht]
  \centering
  \includegraphics[width=\linewidth]{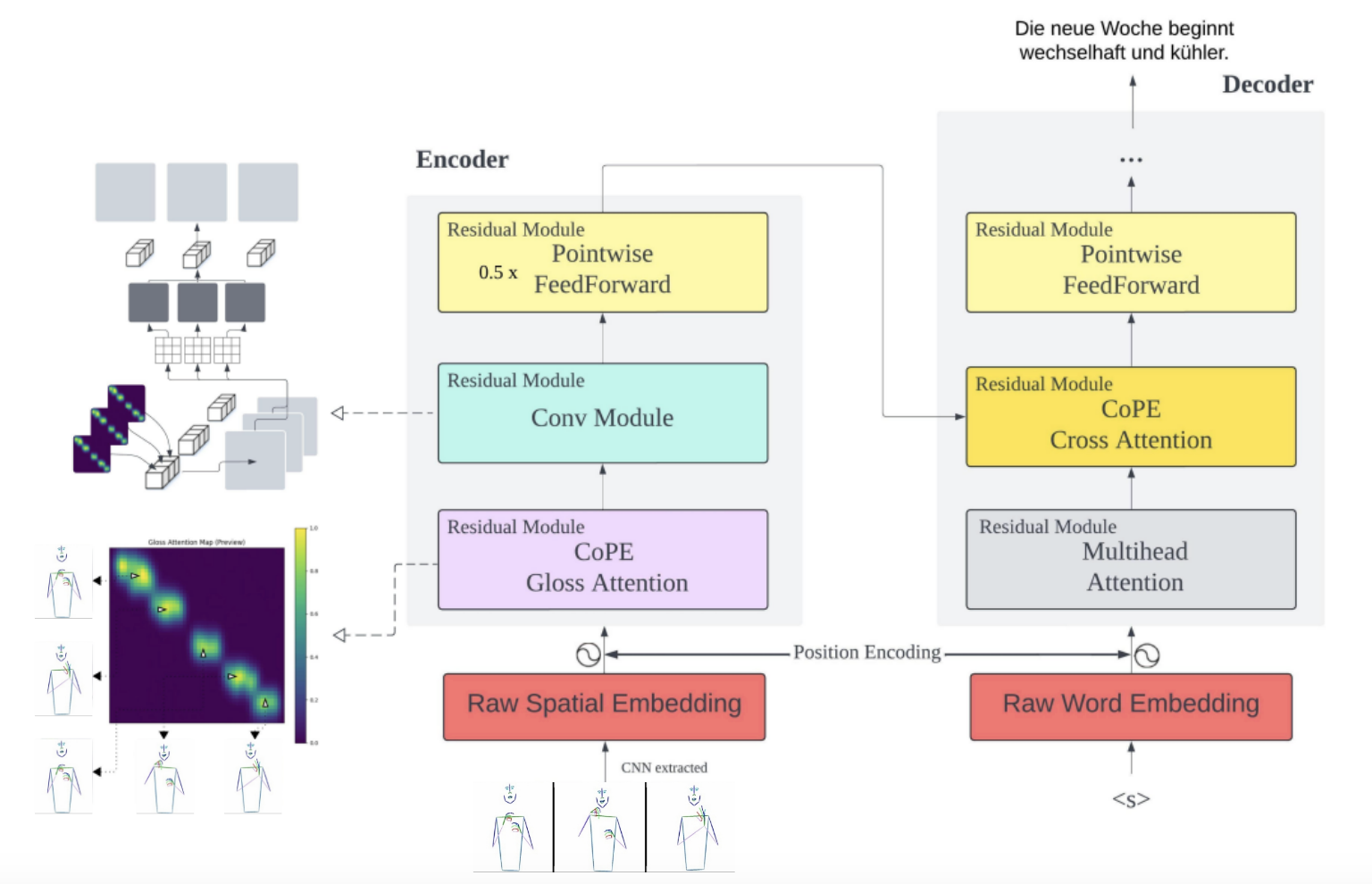}
  \caption{Overview of the adapted \textit{Signformer} architecture (originally from \cite{yang2024signformer}) used in our experiments. Instead of the CNN-extracted frame tokens in the original, each frame's concatenated hand, upper-body, and selected facial landmarks (normalized and linearly projected) feed the encoder directly as pose keypoints.}
  \label{fig:signformer-architecture}
\end{figure}

Our baseline follows the standard Signformer encoder–decoder Transformer architecture \cite{yang2024signformer} (Figure~\ref{fig:signformer-architecture}), adapted to accept skeleton keypoints as input in place of visual features. Instead of spatio-temporal visual embeddings extracted from raw video frames, each input video is represented as a sequence of pose keypoints. Specifically, MediaPipe Holistic \cite{maia2025mediapipe} extracts 2D coordinates for 33 body landmarks, 21 landmarks per hand (left and right), and a facial subset around the mouth and eyebrows. These are concatenated into a per-frame feature vector, producing a time series of pose features that is projected into the model's embedding space through a linear layer before entering the encoder.
This representation is motivated by two practical considerations. First, landmark sequences emphasize articulatory motion rather than appearance, suppressing much of the background, lighting, and camera variation that differs across datasets and signers. Second, self-attention is well suited to continuous signing, where semantically relevant cues may be distributed across long temporal windows and multiple channels such as the hands, face, and upper body. In effect, the model can devote more of its capacity to temporal structure and coordinated movement rather than to reconstructing visual texture.

Using skeleton data substantially reduces input dimensionality and removes much of the irrelevant visual clutter, which can make training and inference more practical for resource-constrained environments \cite{yang2024signformer}. The trade-off is that some information is inevitably lost, including fine-grained appearance cues and subtle gestures that are not well captured by keypoints. Prior work suggests that pose-based approaches may therefore lag behind image-based models on unconstrained translation tasks \cite{zelezny2025exploring}. We treat pose extraction quality as a critical dependency: missing hand tracks, unstable facial landmarks, or brief occlusions can directly degrade the encoder input. In practice, we mitigate these issues with standardized preprocessing and sequence normalization, and we interpret the reported results as comparisons of augmentation policies within a fixed pose-based pipeline. The absence of a heavy visual backbone also simplifies the engineering stack: fewer components must be tuned, memory can be reallocated to longer sequences or larger batches, and preprocessing failures are easier to diagnose at the landmark level. Although our absolute BLEU scores remain below state of the art systems that operate on full video (see Section~\ref{sec:results}), the relative comparisons with and without augmentation remain informative within this setting.

\subsection{LLM-Based Data Augmentation}

To augment the training data, we use GPT-4o as a paraphrase generator. For each original video--sentence pair $(V, T)$, where $T$ is the ground-truth spoken-language translation of the sign video $V$, we generate $N=3$ additional sentences $T_1', T_2', T_3'$ intended to preserve the meaning of $T$. The prompt instructs the model to produce semantically faithful paraphrases in the original target language, preserving propositional content while allowing controlled lexical and syntactic variation. This is desirable because the sign input remains fixed while the decoder sees a broader set of valid target realizations.

The prompt requires a structured JSON response, which makes the pipeline easier to validate automatically and easier to resume after interruptions. The instructions explicitly require the model to preserve tense, register, and propositional content, allowing changes in wording or word order only when the meaning remains intact.

\begin{figure*}[t]
\centering
\includegraphics[width=\linewidth]{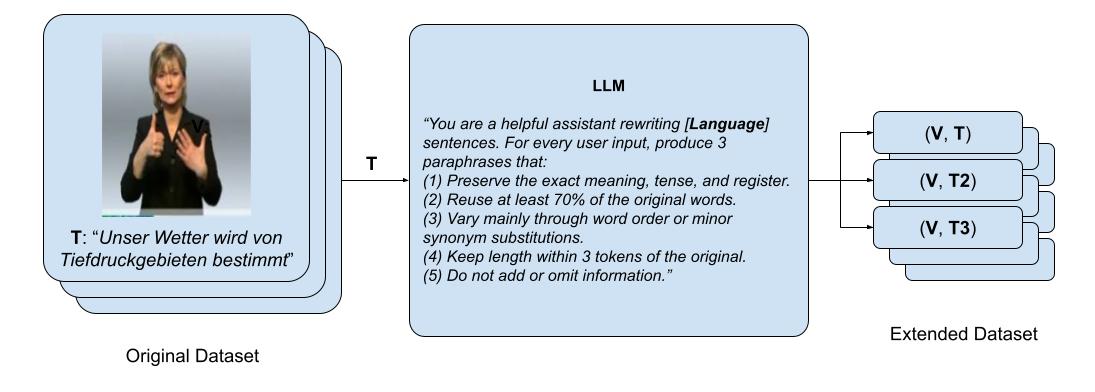}
\caption{Overview of the LLM-augmented SLT pipeline. For each video--text pair $(V,T)$, an LLM generates three paraphrases $(T'_1,T'_2,T'_3)$ that preserve meaning while introducing controlled lexical and syntactic variation. Training then follows two stages: pre-training on the augmented corpus and fine-tuning on the original targets only. At inference time, the model translates directly from sign input to text.}
\label{fig:pipeline}
\end{figure*}

For each original sign video $V$, the resulting training set contains the reference $T$ and three paraphrases $T'_1, T'_2, T'_3$. During training, these are materialized as four separate examples $(V, T)$, $(V, T'_1)$, $(V, T'_2)$, and $(V, T'_3)$, meaning that the same sign input is paired with multiple textual realizations. Figure~\ref{fig:pipeline} summarizes the overall procedure, and Table~\ref{tab:aug-examples} provides representative examples from PHOENIX14T.

Not every generated sentence is retained. To reduce semantic drift and remove trivial copies, we filter candidate paraphrases with four surface-form similarity measures between the original sentence and each variant: character-level Jaccard overlap, word-level Jaccard overlap, normalized Levenshtein similarity, and trigram overlap. Let $\bar{s}(T, T_i')$ denote the mean of these four scores. We keep a variant only if
\[
0.3 \leq \bar{s}(T, T_i') \leq 0.95,
\]
and we discard exact duplicates regardless of the threshold. The lower bound rejects variants that are too dissimilar and therefore more likely to have altered the meaning, while the upper bound removes nearly identical rewrites that contribute little diversity. In practice this lightweight filtering step proved important because LLM outputs occasionally oscillated between overly literal copies and overly free reformulations. The generation pipeline also includes checkpointing so that long batch runs can be resumed without regenerating already accepted samples. In production, outputs were required to return structured JSON so that results could be validated automatically before entering the training set; the full prompt template is shown in Figure~\ref{fig:pipeline}.

\begin{table*}[t]
\centering
\small
\begin{tabular}{@{}p{0.35\linewidth}@{\hspace{1em}}p{0.56\linewidth}@{}}
\hline
\textbf{Original (reference)} & \textbf{LLM paraphrases} \\
\hline
\parbox[t]{\linewidth}{tiefdruckgebiete bestimmen unser wetter\\
\textit{\small low-pressure areas determine our weather}}
&
\parbox[t]{\linewidth}{%
\textbullet\ Unser Wetter wird von Tiefdruckgebieten bestimmt.\\
\textit{\small Our weather is determined by low-pressure areas.}\\[2pt]
\textbullet\ Die Bestimmung unseres Wetters erfolgt durch Tiefdruckgebiete.\\
\textit{\small The determination of our weather is due to low-pressure areas.}}
\\
\hline
\parbox[t]{\linewidth}{auch mit den temperaturen geht es aufwärts\\
\textit{\small the temperatures are also rising}}
&
\parbox[t]{\linewidth}{%
\textbullet\ Auch die Temperaturen steigen an.\\
\textit{\small The temperatures are also increasing.}\\[2pt]
\textbullet\ Die Temperaturen gehen ebenfalls nach oben.\\
\textit{\small The temperatures are also going up.}}
\\
\hline
\parbox[t]{\linewidth}{eine gewitterfront überquert deutschland von west nach ost\\
\textit{\small a thunderstorm front crosses Germany from west to east}}
&
\parbox[t]{\linewidth}{%
\textbullet\ Eine Gewitterfront zieht von Westen nach Osten über Deutschland.\\
\textit{\small A thunderstorm front moves from west to east across Germany.}\\[2pt]
\textbullet\ Von Westen nach Osten überquert eine Gewitterfront Deutschland.\\
\textit{\small From west to east, a thunderstorm front crosses Germany.}}
\\
\hline
\end{tabular}
\caption{Examples of GPT-4o paraphrases paired with original PHOENIX14T training references.}
\label{tab:aug-examples}
\end{table*}

\subsection{Training Schedule}
\label{sec:train-schedule}
We compare two conditions:
\begin{itemize}\setlength{\itemsep}{1pt}
    \item \textbf{Baseline:} the model is trained only on the original, non-augmented training set.

    \item \textbf{+Augmentation:}
    \begin{itemize}
        \item \emph{Stage1:} pre-train on the augmented corpus consisting of the original targets plus three GPT-4o paraphrases per instance.
        \item \emph{Stage2:} fine-tune on the original training set only, so that the decoder is re-aligned with the reference phrasing and less likely to overproduce rare paraphrastic variants. Unless otherwise stated, all hyperparameters remain identical across conditions, and early stopping is performed on the same development set.
    \end{itemize}
\end{itemize}

This two-stage schedule is central to the method. Training exclusively on paraphrastic variants can broaden the decoder distribution, but it can also bias the model toward outputs that are semantically valid yet deviate from the single reference used at evaluation time. The final fine-tuning stage partially counteracts that tendency by re-centering the model on the original corpus style without discarding the broader lexical exposure learned during pre-training. For fairness, both conditions use the same optimization policy, validation protocol, and stopping criterion; the intended difference is only the presence or absence of augmented targets. The training setup follows standard sequence-to-sequence practice: teacher forcing with cross-entropy loss, a warm-up-and-decay learning-rate schedule, label smoothing, and early stopping on validation loss.

\section{Datasets and Evaluation}

We evaluate the proposed approach on three SLT datasets that differ markedly in linguistic diversity, recording conditions, and lexical structure, all of which influence how target-side augmentation behaves.

The \textbf{PHOENIX14T} dataset \cite{camgoz2018neural} contains weather-broadcast recordings in German Sign Language (DGS) paired with German text. It is a real-world corpus with relatively consistent, domain-specific phrasing and limited topic variation. That repetitiveness simplifies part of the translation problem, but the natural recording conditions still introduce visual variation across signers and sessions, yielding a moderately challenging benchmark.

The \textbf{GSL} dataset \cite{adaloglou2020comprehensive}, in contrast, is recorded under controlled laboratory conditions and features a small set of signers repeatedly producing a restricted inventory of predefined sentences. As a result, it has low linguistic and visual variability, high redundancy across samples, and almost no rare target tokens. Models can therefore memorize sentence patterns easily and achieve near-perfect BLEU, but this simplicity also makes the benchmark sensitive to any increase in output variation.

Finally, \textbf{LSA-T} \cite{dalbianco2022lsat} consists of real-world videos collected from diverse sources, with substantial variation in signers, lighting, and signing style. Its naturalistic content and broad Spanish vocabulary make it much harder than the previous two datasets. The high proportion of singleton words and irregular phrasing creates a pronounced long-tail distribution, leading to sparse lexical coverage and very low baseline translation accuracy. This makes LSA-T a useful test bed for assessing whether augmentation can help under severe data sparsity.

Table~\ref{tab:dataset-stats} quantifies the key differences across these three settings.

For all datasets, videos are preprocessed with MediaPipe to extract pose sequences, as described above. Coordinate values are normalized per frame sequence and frame rates are standardized for model input following steps similar to \cite{zelezny2023poseSLT}, while the textual side is lowercased and tokenized consistently across corpora. Translation quality is measured with case-insensitive BLEU-4 \cite{papineni2002bleu} on the official test splits.

\begin{table*}[t]
\centering
\small
\begin{tabular}{lccc}
\hline
\textbf{Statistic} & \textbf{PHOENIX14T (DGS)} & \textbf{GSL} & \textbf{LSA-T (LSA)} \\
\hline
Language (target) & German & Greek & Spanish \\
Sign language & DGS & GSL & LSA \\
Real-world footage & Yes & No & Yes \\
No. of signers & 9 & 7 & 103 \\
Duration [h] & 10.71 & 9.51 & 21.78 \\
Samples (clips) & 7{,}096 & 10{,}295 & 8{,}459 \\
\hline
Unique sentences & 5{,}672 & 331 & 8{,}102 \\
\% unique sentences & 79.93\% & 3.21\% & 95.79\% \\
Vocabulary size (types) & 2{,}887 & N/A & 14{,}239 \\
Singletons (types with count{=}1) & 1{,}077 & 0 & 7{,}150 \\
\% singletons & 37.3\% & 0\% & 50.21\% \\
\hline
Resolution & 210$\times$260 & 848$\times$480 & 1920$\times$1080 \\
FPS & 25 & 30 & 30 \\
\hline
\end{tabular}
\caption{Corpus statistics for the three datasets used in the main experiments. The lower block highlights lexical properties related to long-tail behavior.}
\label{tab:dataset-stats}
\end{table*}

\section{Results and Analysis}
\label{sec:results}

\subsection{Main Three-Dataset Benchmark}

Table~\ref{tab:bleu-results} reports BLEU-4 on the test sets for the baseline model and the two-stage \textbf{+Augmentation} configuration.

\begin{table*}[t]
\centering
\begin{tabular}{lcc}
\hline
\textbf{Dataset} & \textbf{Baseline (BLEU-4)} & \textbf{+Augmentation (BLEU-4)} \\
\hline
PHOENIX14T (DGS) & 9.56 & 10.33 \\
GSL (Greek) & 94.38 & 92.22 \\
LSA (Spanish) & 1.18 & 1.19 \\
\hline
\end{tabular}
\caption{Test BLEU-4 for baseline training and LLM-augmented training on the three evaluation datasets.}
\label{tab:bleu-results}
\end{table*}

\paragraph{Overall trends.} PHOENIX14T shows an improvement of +0.77 BLEU. In a moderately rich yet still formulaic domain, exposure to paraphrastic reorderings appears to help the decoder generalize beyond memorized templates, while the final fine-tuning stage helps keep the output close to the evaluation style. By contrast, the GSL benchmark starts from an exceptionally high baseline of 94.38 BLEU, reflecting substantial overlap and low linguistic variability between training and test. In that near-saturated setting, augmentation slightly reduces performance to 92.22 BLEU: the decoder learns alternative phrasings that remain semantically valid but do not exactly match the single reference, and fine-tuning does not completely suppress those variants. Finally, the LSA-T setting remains extremely low in both conditions, at roughly 1.2 BLEU. Because the paraphrasing prompt deliberately preserves the same rare content words, target-side augmentation does not address the underlying problem of severe data sparsity on the sign side.

\paragraph{Data characteristics matter.} The utility of LLM paraphrasing is closely tied to vocabulary breadth and to the prevalence of infrequent tokens. When a dataset provides enough lexical variety, as in PHOENIX14T, paraphrastic exposure can help the model tolerate alternative word orderings and light lexical substitutions at test time. When the task is unusually simple, as in the GSL benchmark, greater output diversity can reduce single-reference BLEU even if the underlying meaning is preserved. When the vocabulary is extremely sparse, as in the LSA-T setting, paraphrasing the target text alone does not solve the central coverage problem: many content words and corresponding sign patterns remain too rare for the decoder to learn robustly.

\paragraph{Why BLEU can understate gains.} Target-side augmentation creates a methodological mismatch at evaluation time. During training, the model is encouraged to treat multiple textual realizations of the same signed content as acceptable. BLEU, however, rewards overlap with only one reference sentence. For a weather forecast, training may expose the decoder to alternatives such as ``morgen wird es regnen,'' ``es regnet morgen,'' and ``f\"ur morgen ist Regen vorhergesagt.'' If the model later produces another semantically correct variant that differs lexically from the single test reference, BLEU will penalize it despite preserving the meaning. The objective encouraged by augmentation is therefore broader than the one measured by single-reference lexical overlap, which helps explain why a modest BLEU gain can still correspond to a more meaningful improvement in semantic robustness.

\subsection{Semantic Evaluation via LLM-as-a-Judge}

The lexical-overlap mismatch discussed above motivates a semantic evaluation of the PHOENIX14T and GSL runs from the previous section. An LLM judge scores translations along semantic fidelity and linguistic quality, counts \emph{fluent but wrong} outputs, and performs direct pairwise comparisons between baseline and augmented hypotheses.

Recent literature supports this direction. Strong LLM judges have shown high agreement with human preferences on open-ended generation benchmarks \cite{zheng2023judging}, and LLM-based translation evaluators have achieved competitive or state-of-the-art correlation with human judgments in machine translation \cite{kocmi-federmann-2023-large}. More structured prompting strategies such as G-Eval \cite{liu-etal-2023-g} further suggest that dimension-specific scoring can make automated judging more interpretable. At a lighter-weight end of the spectrum, embedding-based methods such as Sentence-BERT \cite{reimers-gurevych-2019-sentence} provide an additional semantic signal between pure lexical overlap and full LLM judging.

\begin{table}[t]
\centering
\small
\begin{tabular}{lccc}
\hline
\textbf{Dataset} & \textbf{Baseline} & \textbf{Augmented} & \textbf{Change} \\
\hline
PHOENIX14T & 2.51 & 3.65 & +45.0\% \\
GSL        & 7.72 & 8.77 & +13.6\% \\
\hline
\end{tabular}
\caption{LLM-as-a-Judge semantic fidelity scores (GPT-5.2) for baseline and augmented models.}
\label{tab:llm-judge-fidelity}
\end{table}

\paragraph{Results.} Table~\ref{tab:llm-judge-fidelity} reports semantic fidelity on both datasets. On PHOENIX14T, augmentation raises fidelity from 2.51 to 3.65 (+45\%). Additional PHOENIX14T metrics reinforce this: the rate of fluent-but-semantically-incorrect translations drops from 54.8\% to 35.5\% ($-19.3$ pp), and pairwise preference judgments favor the augmented model in 52.9\% of comparisons versus 13.1\% for the baseline. On GSL, where the baseline was already strong, fidelity still improves from 7.72 to 8.77 (+13.6\%), confirming semantic gains in a regime where lexical overlap metrics are near-saturated. Together, these results confirm that target-side augmentation improves semantic quality beyond what BLEU alone captures.

\paragraph{Limitations of semantic judging.} Any LLM-based evaluator must be used cautiously. In our experiments, GPT-5.2 served as the judge while GPT-4o generated the augmented training targets. These are architecturally distinct (GPT-5.2 is a reasoning model), which reduces but does not eliminate this concern: shared training lineage may still introduce subtle stylistic alignment between generator and evaluator. For that reason, semantic judging should be treated as complementary evidence and ideally validated with human assessment or at least triangulated with simpler automatic signals. This concern is particularly relevant in SLT, where small lexical differences can be acceptable while subtle semantic errors remain difficult to detect automatically.

\section{Conclusion}
Target-side LLM augmentation is not uniformly beneficial for SLT: its value depends on where a corpus sits along the spectrum from formulaic to long-tail sparse. On PHOENIX14T, paraphrastic exposure improves BLEU and reduces semantically incorrect fluent outputs. On GSL, the near-perfect baseline leaves no room for lexical variation to help and single-reference BLEU penalizes it. On LSA-T, target-side rewriting cannot address sign-side sparsity. The central lesson is that augmentation works when a dataset has enough lexical breadth for variation to improve generalization, but not so little diversity that alternative valid phrasings are penalized at evaluation time.

The semantic evaluation confirms that gains from augmentation exceed what BLEU alone captures, while the architectural gap between GPT-4o (augmentation) and GPT-5.2 (judge, a reasoning model) reduces but does not eliminate the risk of shared stylistic bias; future work should validate with human judgments or cross-family evaluators.

Promising directions include combining target-text augmentation with sign-side augmentation so that linguistic and motion variation are learned jointly; evaluating with multiple references or human judgments, where semantically valid paraphrases are less likely to be penalized; and replacing GPT-4o with open-source LLMs or task-specific paraphrasers to clarify whether the gains stem from the augmentation principle or from the specific generator.

{
    \small
    \bibliographystyle{ieeenat_fullname}
    \bibliography{custom}
}


\end{document}